%% file: root.tex
\documentclass[letterpaper, 10pt, conference]{ieeeconf}
\IEEEoverridecommandlockouts
\input{preamble}

\title{\LARGE \bf
  Motions in Microseconds via Vectorized Sampling-Based Planning
}

\author{Wil Thomason\textsuperscript{\dag}, Zachary Kingston\textsuperscript{\dag}, and Lydia E. Kavraki
  \thanks{
    \dag~Equal contribution.
    All authors are affiliated with the Department of Computer Science, Rice University, Houston TX, USA
    {\tt\small \{wbthomason, zak, kavraki\}@rice.edu}.
    This work was supported in part by NSF RI 2008720, NSF ITR 2127309 for the Computing Research Association CIFellows Project, and Rice University Funds.
}}

\begin{document}
\maketitle
\thispagestyle{empty}
\pagestyle{empty}

\begin{abstract}

Modern sampling-based motion planning algorithms typically take between hundreds of milliseconds to dozens of seconds to find collision-free motions for high degree-of-freedom problems.
This paper presents performance improvements of more than 500x over the state-of-the-art, bringing planning times into the range of microseconds and solution rates into the range of kilohertz, without specialized hardware.
Our key insight is how to exploit fine-grained parallelism within sampling-based planners, providing generality-preserving algorithmic improvements to any such planner and significantly accelerating critical subroutines, such as forward kinematics and collision checking.
We demonstrate our approach over a diverse set of challenging, realistic problems for complex robots ranging from 7 to 14 degrees-of-freedom.
Moreover, we show that our approach does not require high-power hardware by also evaluating on a low-power single-board computer.
The planning speeds demonstrated are fast enough to reside in the range of control frequencies and open up new avenues of motion planning research.
\end{abstract}

\section{Introduction}\label{sec:introduction}
\input{sections/introduction}

\section{Related Work}\label{sec:related-work}
\input{sections/relatedwork}

\section{Method}\label{sec:method}
\input{sections/method}

\section{Experiments}\label{sec:experiments}
\input{sections/experiments}

\section{Discussion}\label{sec:discussion}
\input{sections/discussion}

\section*{Acknowledgements}
The authors would like to thank Mark Moll for help evaluating MoveIt, Sofia Paola Medina-Chica for
 \arm development, and Stefan Bukorovic for collision primitive development.

\printbibliography{}

\end{document}

%% file: preamble.tex
\usepackage[utf8]{inputenc}

\usepackage{comment}
\usepackage{xspace}
\usepackage{graphicx}
\usepackage{overpic}
\usepackage{svg}
\usepackage{xcolor}
\definecolor{gg}{HTML}{2b3b5e}
\definecolor{Gray}{gray}{0.95}

\definecolor{setgreen}{HTML}{66c2a5}
\definecolor{setorange}{HTML}{fc8d62}
\definecolor{setblue}{HTML}{8da0cb}
\definecolor{setpurp}{HTML}{e78ac3}

\usepackage{tabularx}
\usepackage{multirow}
\usepackage{booktabs}
\usepackage{colortbl}
\usepackage{siunitx}

\usepackage{etoolbox}
\makeatletter
\patchcmd{\@makecaption}
{\scshape}
{}
{}
{}
\makeatother

\let\labelindent\relax
\usepackage[inline]{enumitem}

\usepackage{amsmath}
\usepackage{amssymb}
\usepackage{amsthm}

\newtheoremstyle{main}
{1em}                                                %
{1em}                                                %
{\itshape}                                        %
{0pt}                                                %
{\scshape}                                           %
{\\*}                                                %
{2pt}                                                %
{\thmname{#1}\thmnumber{ #2}: \thmnote{\itshape #3}} %

\usepackage[linesnumbered,ruled,noend]{algorithm2e}

\usepackage[
	activate   = {true},
	protrusion = false,
	expansion  = true,
	kerning    = true,
	spacing    = true,
	tracking   = false,
	auto       = true,
	selected   = true,
	factor     = 1000,
	stretch    = 20,
	shrink     = 20,
]{microtype}

\usepackage[english=american]{csquotes}
\MakeOuterQuote{"}
\usepackage[
	maxbibnames=99,
	maxcitenames=2,
	natbib=true,
	style=numeric-comp,
	backend=biber,
	sorting=none,
	giveninits=true,
	url=false,
	doi=false,
	eprint=true,
	isbn=false,
]{biblatex}

\makeatletter
\let\NAT@parse\undefined
\makeatother
\usepackage[pdfa,colorlinks,bookmarksopen,bookmarksnumbered,allcolors=gg]{hyperref}
\usepackage[english]{babel}

\usepackage[nameinlink,capitalise]{cleveref}
\crefname{line}{line}{lines}
\crefname{figure}{Fig.}{Figs.}
\Crefname{figure}{Fig.}{Figs.}
\crefname{equation}{Eq.}{Eqs.}
\Crefname{equation}{Eq.}{Eqs.}
\crefname{section}{Sec.}{Secs.}
\Crefname{section}{Sec.}{Secs.}
\crefname{definition}{Def.}{Defs.}
\Crefname{definition}{Def.}{Defs.}
\crefname{algorithm}{Alg.}{Algs.}
\Crefname{algorithm}{Alg.}{Algs.}
\crefname{assumption}{Asm.}{Asms.}
\Crefname{assumption}{Asm.}{Asms.}
\crefname{subassumption}{Asm.}{Asms.}
\Crefname{subassumption}{Asm.}{Asms.}

\newcommand{\mf}[1]{\mbox{\cref{#1}}\xspace}

\usepackage{flushend}

\graphicspath{ {./figures} }

\newcommand{\ompl}{\textsc{ompl}\xspace}
\newcommand{\ros}{\textsc{ros}\xspace}
\newcommand{\sse}{\textsc{sse}\xspace}
\newcommand{\arm}{\textsc{arm}\xspace}

\newcommand{\avxtwo}{\textsc{avx$2$}\xspace}
\newcommand{\avxthree}{\textsc{avx-$512$}\xspace}
\newcommand{\simd}{\textsc{simd}\xspace}
\newcommand{\cpu}{\textsc{cpu}\xspace}
\newcommand{\gpu}{\textsc{gpu}\xspace}
\newcommand{\amd}{\textsc{amd}\xspace}
\newcommand{\cpus}{\textsc{cpu}s\xspace}
\newcommand{\gpus}{\textsc{gpu}s\xspace}
\newcommand{\sbmp}{\textsc{sbmp}\xspace}
\newcommand{\sbmps}{\textsc{sbmp}s\xspace}
\newcommand{\rrt}{\textsc{rrt}\xspace}
\newcommand{\rrtconnect}{\textsc{rrt}-Connect\xspace}
\newcommand{\rrtstar}{\textsc{rrt}$^*$\xspace}
\newcommand{\prm}{\textsc{prm}\xspace}
\newcommand{\urdf}{\textsc{urdf}\xspace}

\newcommand{\cuda}{\textsc{cuda}\xspace}

\newcommand{\asao}{\textsc{asao}\xspace}
\newcommand{\eg}{\emph{e.g.},\xspace}
\newcommand{\ie}{\emph{i.e.},\xspace}
\newcommand{\dof}{\textsc{d\scalebox{.8}{o}f}\xspace}

\newcommand{\aos}{\textsc{a\scalebox{.8}{o}s}\xspace}
\newcommand{\soa}{\textsc{s\scalebox{.8}{o}a}\xspace}
\newcommand{\mpc}{\textsc{mpc}\xspace}
\newcommand{\asic}{\textsc{asic}\xspace}
\newcommand{\fpga}{\textsc{fpga}\xspace}
\newcommand{\pspace}{\textsc{pspace}\xspace}
\newcommand{\fk}{\textsc{fk}\xspace}
\newcommand{\nn}{\textsc{nn}\xspace}
\newcommand{\cc}{\textsc{cc}\xspace}

\newcommand{\moveit}{{\color{setorange}\textbf{MoveIt/\ompl}}\xspace}
\newcommand{\grapeshot}{{\color{setgreen}\textbf{PyBullet/\ompl}}\xspace}
\newcommand{\vamp}{{\color{setblue}\textbf{VAMP}}\xspace}
\newcommand{\vamparm}{{\color{setpurp}\textbf{VAMP (\arm)}}\xspace}

\usepackage{mathtools}

\usepackage{siunitx}

%% file: sections/introduction.tex
High degree-of-freedom (\dof) robots rely on \emph{motion planning} to move in complex workspaces, either using sampling-based approximations~\cite{Kuffner2000,Choset2005,LaValle2006,Kavraki2016,Kavraki1996} or numerical optimization~\cite{Zucker2013,Schulman2014}.
These planners are general and can solve realistic, challenging problems in hundreds of milliseconds to dozens of seconds on consumer \cpus{}.
However, this level of performance falls short---it is too slow for reactive operation in evolving environments and hampers algorithms for higher-level autonomy such as integrated task and motion planning.

A large literature accelerates motion planning with \emph{coarse-grained} (\eg thread or process-level) parallelism~\cite{agboh_realtime_online_2018,amato_probabilistic_roadmap_1999,bialkowski_massively_parallelizing_2011,jacobs_scalable_method_2012,park_realtime_optimizationbased_2013}, but these methods have seen relatively little uptake in practice, as their performance gains do not justify the added complexity.
More recent work~\cite{bhardwaj_storm_integrated_2021,sundaralingam2023curobo,fishman2023mpinets} uses \gpu{}-based parallelism to improve performance, but at the cost of communication overhead, algorithmic limitations, and the additional expense and power consumption of \gpu{} hardware.
In general, the field has come to believe that sampling-based motion planner (\sbmp{}) primitives (\eg checking if motion between two states is valid) are either inherently serial, cannot be accelerated without specialized hardware, or cannot be parallelized without paying a greater cost than the parallelism saves.

We refute this belief and show several orders of magnitude performance improvement over the state-of-the-art (more than 500x faster) by contributing insights into \emph{fine-grained} parallelism and work-ordering in \sbmps.
Our core insight is the use of \emph{vector-oriented} state representations and planning primitives (\eg forward kinematics and collision checking), which enable fine interleaving of parallel and serial operation. %
Crucially, we use ``Single Instruction/Multiple Data'' (\simd) instructions to execute these primitives with high throughput and low latency on ubiquitous consumer \cpus, accelerating \emph{almost any} \sbmp for "free".
These insights let us plan high-quality paths at reactive speeds (\eg a median time of \SI{40}{\micro\second} for the 7~\dof Panda over the MotionBenchMaker~\cite{chamzas2021mbm} dataset, \ie \SI{25}{\kilo\hertz}---see~\cref{tbl:results}) on a single \cpu core.\looseness=-1 %

Our method significantly outperforms standard implementations~\cite{Sucan2012} of state-of-the-art \sbmps on both desktop and low-power single board computers.
Moreover, this work will enhance any work that uses motion planning, and our perspective on vector-oriented planning primitives extends beyond \cpu{} \simd{} instructions to other, similar parallelism models, \eg \gpus{}. %
The planning speeds demonstrated blur the line between planning and control, and give cause to re-evaluate assumptions about robot motion.

%% file: sections/relatedwork.tex
Motion planning is \pspace-complete~\cite{reif1979complexity, canny1988complexity} in general, but sampling-based motion planners (\sbmps) empirically can efficiently solve challenging problems.
\sbmp requires computationally expensive subroutines, \eg forward kinematics (\fk), collision checking (\cc), and nearest neighbor (\nn) search~\cite{LaValle2006}.
\cc dominates computational cost in \sbmp (noted by many, \eg~\cite{bialkowski_massively_parallelizing_2011}); however, \nn search may dominate in high-dimensions or with large numbers of states~\cite{kleinbort2016collision}.
\fk{} is used by \cc and contributes to its cost.
For efficiency, \sbmps must use fast \cc and avoid as much \fk/\cc as possible---as such, in this work, we target \fk and \cc performance.

Most \cc algorithms use a \emph{broadphase} to approximate the set of possible collisions, and a \emph{narrowphase} to find exact collisions from the approximate set~\cite{Ericson2004}.
The broadphase is vital for performant \cc and can be accelerated with \gpu~\cite{Lauterbach2009} or \simd~\cite{Tan2019} parallelism.
Efficient, general \cc libraries (\eg~\cite{Pan2012a, Coumans2016}) offer many options for these phases.
Given its impact on \sbmp performance, many works have reduced \cc{} effort via \eg heuristic ordering~\cite{Sanchez2003}, lazy checking~\cite{Bohlin2000,Haghtalab2018}, checking only likely-valid edges~\cite{Nielsen2000,Choudhury2017}, etc.
Recent work has learned a combined \fk{}/\cc "primitive"~\cite{Das2020,Danielczuk2021,Murali2023}; other work has learned distance-to-collision functions~\cite{Rakita2018,koptev_neural_joint_2023}.

\subsection{Parallelism in Motion Planning}\label{sec:related:parallel}

Parallelized planners have been sought since the advent of motion planning (\eg~\citet{Barraquand1990,Henrich1997}).
We broadly categorize parallelism in motion planning as either \emph{coarse-grained} or \emph{fine-grained}.
In our use, coarse-grained refers to parallelism at the level of subroutines or planner components, such as running many planners in parallel, or running \cc in a separate thread.
Fine-grained refers to parallelism at the level of primitive operations, such as checking several states for collisions simultaneously in the same thread, without architectural changes.

Parallelism in \sbmp is typically coarse-grained, \eg simply running independent planners in parallel.
This improves average-case performance~\cite{Wedge2008}; the set of solutions can also be hybridized together to improve plan quality~\cite{Raveh2010}.
Early work (\eg~\citet{amato_probabilistic_roadmap_1999}) observed that roadmap-based planners (\eg \prm~\cite{Kavraki1996}) are amenable to coarse-grained parallelism.
Parallel \sbmp has also been achieved by constructing a forest of planning trees~\cite{Plaku2005}, potentially in distinct regions of the search space~\cite{jacobs_scalable_method_2012}, and by parallelizing components of the \rrtstar \asao planner~\cite{Xiao2017,Ichnowski2012}.
These methods typically offer sub-linear (in the degree of parallelism) performance improvement (with some exceptions~\cite{Ichnowski2012}) due to the synchronization overhead and architectural complexity required.
Still other work uses coarse-grained parallelism in graph search used in roadmap- or search-based planning, both on \cpus~\cite{mukherjee_epa_se_2022,mukherjee_mplp_massively_2022} and \gpus~\cite{zhou_massively_parallel,fukunaga_survey_parallel_2017,fukunaga_parallel_statespace_2018}.
In contrast to the work discussed above, this work investigates a novel approach to \emph{fine-grained} parallelism in \sbmps, which is understudied in the field.\looseness=-1

Recently, there has been significant interest in applying \gpus more broadly to motion planning.
The most successful approaches include parallelized sampling-based \mpc~\cite{bhardwaj_storm_integrated_2021}, parallel particle-based optimization seeded by a partially parallelized \rrt-like planner~\cite{sundaralingam2023curobo}, and end-to-end learning of a neural local control policy from a dataset of motion plans~\cite{fishman2023mpinets}.
Earlier work also investigated \gpu-parallelized \cc~\cite{bialkowski_massively_parallelizing_2011,Pan2012}.
Although these methods show promising performance, they require powerful \gpus for efficiency and impose the overhead of moving data between the \gpu and \cpu.

\subsection{Hardware-Accelerated Motion Planning}

Hardware acceleration is crucial for our proposed vector-based approach to \sbmp.
We use \simd instructions, a feature ubiquitous on consumer \cpus\footnote{Our primary implementation uses \avxtwo, which has been broadly available on Intel and AMD \cpus since 2013. Lower-width \simd instruction sets such as \sse have been available since 1999~\cite{Intel2011}.}.
Hardware acceleration has long been used for motion planning, with particular focus on accelerating \cc~\cite{Kameyama1992}.
Some work (\eg~\citet{Ichnowski2019}'s compile-time specialized \sbmp, or~\citet{Carpentier2019}'s statically dispatched, precompiled dynamics algorithms) implicitly exploits hardware acceleration by creating "machine sympathetic" implementations---code that enables compilers, etc. to better exploit hardware capabilities.

Many modern robotics algorithms are accelerated by \gpus, \eg using \cuda.
\gpu acceleration has been applied to \sbmps~\cite{ichter_group_marching_2017, lawson_gpu_parallelization_2020, sundaralingam2023curobo}, \mpc~\cite{hyatt_parameterized_gpuparallelized_2020, bhardwaj_storm_integrated_2021} and trajectory optimization~\cite{park_realtime_optimizationbased_2013, sundaralingam2023curobo}.
Recently, \asic- or \fpga-based accelerators~\cite{wan2021fpga} have been proposed, \eg to validate an entire roadmap at once with an \fpga~\cite{murray2016chip, murray2016micro, murray2019thesis} or as an external \cc accelerator~\cite{shah_energyefficient_realtime_2023}.
\citet{neuman_robomorphic_computing_2021} investigated ``robomorphic'' computing with specialized accelerators for common robotics algorithms.

However, \gpus and other accelerators come at a cost: there is latency in communicating with the device, there are restrictions on the types of algorithms that can be applied on specialized hardware, and often there is a relatively high cost to send data back and forth~\cite{shah_energyefficient_realtime_2023,werkhoven_cpugpu_2014}.
Our approach uses native \simd instructions on the \cpu, inflicting at worst a slight overhead penalty\footnote{\cpus may downclock when using \simd instructions---there is also a cost to move data in and out of vector registers.} to achieve large performance gains.

%% file: sections/method.tex
Most \sbmps can be decomposed into a handful of "primitive" operations (see Ch. 7 in~\citet{Choset2005}).
Algorithms typically use (approximate) nearest-neighbors (\nn) to find nearby states, a state validity function (\eg checking for collisions), a local planner or steering function to grow edges between states, and an edge validity function to check these edges.
Validity functions usually require forward kinematics (\fk) to compute the poses of the robot's links in its workspace from a configuration.

We lift a selection of these primitives---\fk{} and state/edge validity checking---to operate over \emph{vectors} of states in parallel.
This lifting immediately accelerates the primitive operations by multiplying their throughput.
More importantly, shifting perspective to vector primitives
\begin{enumerate*}[label=(\arabic*)]
	\item admits low-overhead parallelism that cooperates with sequential code, and
	\item reveals beneficial algorithmic changes to the primitives based on insights about their specific uses in \sbmp
\end{enumerate*}.
This perspective allows us to exploit ubiquitous hardware parallelism via \simd{} instructions, resulting in highly efficient implementations of our vector primitives.
Finally, by focusing on primitives common across \sbmps, we improve the performance of \emph{almost any} \sbmp without requiring significant algorithmic changes.\looseness=-1 %

\subsection{Vectorized Motion Planning}\label{sec:method.vectorized-computation}
"Vectorized" is an overloaded term; we use it in the \simd sense, where a "vector" is a fixed-length set of values with the same scalar datatype (\eg floating-point numbers) and a "vectorized operation" is an operation that transforms all values in a vector independently, in parallel.

This parallelism model is similar to \gpu{} computing (and our lifted vector operations may benefit \gpu-based planners), but with a few key differences.
We focus on \cpu-based \simd{} parallelism---our algorithms run on any modern computer, even those without a \gpu.
This increases applicability and decreases both the barrier to entry and the power consumption of our technique.
\cpu-based \simd{} parallelism is also better-suited to the opportunities for parallelism in \sbmp: it has significantly lower overhead than \gpu{}-, thread-, or process-based parallelism\footnote{While modern hardware is quite parallelism-performant, there is still non-negligible overhead (\eg \gpu-\cpu communication latency) in the tens of microseconds, which can easily add up into the milliseconds.} and is amenable both to fine-grained interleaving of parallel and sequential code and to efficient computation for relatively small workloads.

Exploiting \simd instructions requires careful consideration of data structure and algorithm design, often involving unconventional memory layouts to ensure adequate \emph{data parallelism}. %
Our approach addresses this challenge through a novel Struct-of-Arrays (\soa) memory layout for \fk and \cc.
This choice enables seamless exploitation of data parallelism, allowing us to pose and check multiple configurations for collision in parallel.
The \soa layout stands in contrast to the more common Array-of-Structs (\aos) layout (illustrated in~\mf{fig:aos_soa}), which is less favorable for \simd approaches as it causes memory access patterns that slow access to values. %

Many \sbmp algorithms and subroutines make heavy use of conditional branching, inimical to parallel code.
However, as \cpu-based \simd{} parallelism allows easy interleaving of parallel and sequential code, and as our subroutines have reduced branching, we are able to sidestep this problem more easily compared to other forms of parallelism\footnote{We also benefit from advances in modern hardware, which have produced branch predictors and fused "test-and-branch" instructions that are highly performant on the limited set of branches we retain.}.
These properties mean that our algorithm implementations, despite being parallelized, are close to "standard" algorithms---there is no explicit synchronization or communication code, etc.

\begin{figure}
	\centering
	\includeinkscape[width=0.75\linewidth]{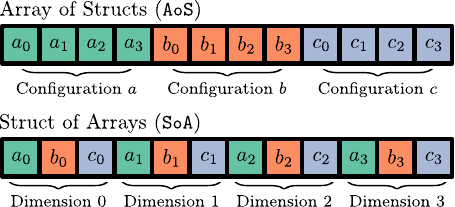}
	\caption{Struct-of-Arrays (\soa) and Array-of-Structs (\aos) memory layout.
		Here, there are three configurations $a, b,$ and $c$, each with four dimensions.
		\aos is the more ``natural'' layout of memory, but hard to exploit with \simd.
	}\label{fig:aos_soa}
\end{figure}

The overhead of conventional mechanisms (\eg threads) limits naive parallelization of \sbmp---reducing this overhead is especially challenging as most \sbmp algorithms rapidly alternate between expensive, parallelization-friendly subroutines (\eg \cc) and code that relies on these subroutines but is not itself easily parallelized (\eg graph search).
Fortunately, the relatively low overhead of \cpu{} \simd-based parallelism---and the specific nature of this overhead, which modern compilers excel at reducing---means that our vector-oriented primitives (the expensive subroutines) can be efficiently interleaved within a \sbmp.
This property may suffice to accelerate \sbmp by, \eg{} reducing the cost of a single collision check.
However, by using \soa memory layouts for our vector-oriented operations, we can not only perform the computation required for \sbmp in parallel, but also exploit motion-planning-specific independence patterns in this computation to \emph{intelligently order} the requisite operations to improve overall performance.

\subsection{Vectorized Forward Kinematics}
Vectorized \fk{} is necessary to pose batches of states in parallel for subsequent parallel \cc; sequentially posing each element in a batch introduces a bottleneck that reduces overall throughput.
Naive vectorization of \fk{} attempts to compute the poses for a \emph{single} configuration faster---we instead choose a less conventional use of vectorization: carrying out each operation in sequence, but on \emph{multiple configurations} simultaneously.

\fk{} implementations commonly use dynamic branching and joint-type polymorphism to compute transforms between links (\eg, \textsc{kdl}~\cite{bruyninckx2001open}).
This structure is difficult for compilers to optimize and has spurious data dependencies between link transforms, which decreases throughput and causes slower operations across the entire vector of configurations.
These dependencies arise as the compiler cannot determine if poses later in the kinematic tree depend on earlier poses (or, better still, on components of these poses).
Even naively vectorizing \fk{} for multiple configurations requires vector configuration and pose data structures, and use of vector operations, which are tedious to manually implement.

We overcome these challenges with a novel \emph{tracing compiler} for robot kinematics.
This compiler takes in standard Universal Robot Description Format (\urdf) files and \emph{traces} the operations of arbitrary functions of the robot's kinematics (\eg \fk).
It uses this trace to automatically generate
\begin{enumerate*}[label=(\arabic*)]
	\item a vector configuration structure representing a batch of configurations and
	\item the minimal set of operations required to compute the traced function
\end{enumerate*}.
This latter output constitutes an "unrolled" \fk{} loop that avoids branching and spurious data dependencies, allowing an optimizing compiler to generate faster machine code.
Further, our tracing compiler applies optimizations to reduce the operations required, \eg constant folding, algebraic simplification, removing redundant negations, etc.
This use of automatic code generation creates hyper-specialized vector-lifted \fk{} without loss of generality, as the tracing compiler itself is general.
We note other techniques for efficient \fk{} by, \eg \citet{Carpentier2019}, which uses static polymorphism and the Curiously Recurring Template Pattern (\textsc{crtp}) for compile-time optimized \fk{} routines.
In contrast, our tracing compiler, by merit of tracking the precise operations (\eg the multiplies, sines, etc. from input configuration to output pose), outputs ``straightline'' code that removes operations that are not necessarily detectable at compile-time with \textsc{crtp}.

\subsection{Vectorized Collision Checking}

Existing approaches to \cc (\eg broadphases, narrowphase triangle mesh collision algorithms~\cite{gilbert_fast_procedure_1987}) are optimized for checking a single configuration.
Fully vectorizing these approaches is challenging.
We instead draw inspiration from classical work on simplified representations of robots and obstacles~\cite{Bradshaw2004,OSullivan1999} to automatically generate collision geometry from meshes using primitives (\ie spheres, cylinders, and cuboids).
Sphere-based representations are common in the trajectory optimization literature~\cite{Schulman2014,fishman2023mpinets}.

By representing the robot and environment as geometric primitives we can vectorize intersection tests between pairs of such primitives.
We check batches of robot poses for self-collision and environment collision in parallel and reject the whole batch if any collide. %
Surprisingly, we see that this narrowphase-only approach (due to its reduced branching) can be highly efficient even in complex environments.
Further, using spheres to represent the robot's geometry synergizes with our tracing compiler for \fk---as we only compute the position of each sphere (rather than a full $SE(3)$ pose), the compiler skips a large number of irrelevant operations.

\subsection{Vectorized Motion Validation}

\begin{figure}
	\centering
	\includeinkscape[width=0.8\linewidth]{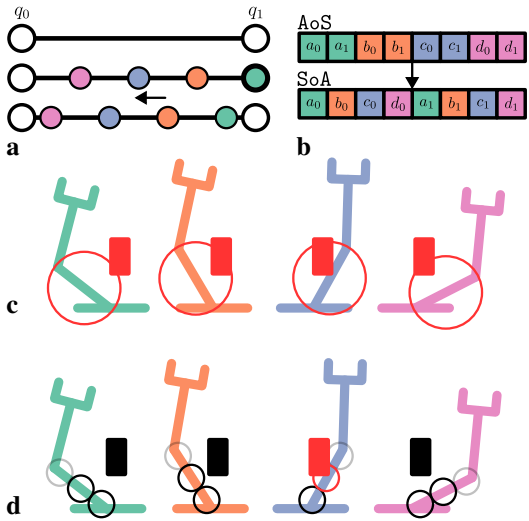}
	\caption{Illustration of the ``raked'' motion validator for a two-link mechanism.
    \textbf{a)} The rake consists of evenly spaced configurations (here, $n = 4$), which are ``raked'' backwards to achieve sufficient resolution.
    \textbf{b)} These configurations are computed in \soa form from an initial \aos layout, then checked in parallel.
    \textbf{c)} Spherical approximations of collision geometry are checked in parallel.
    A hierarchy of spheres is used to avoid unnecessary checks.
    \textbf{d)} When any collision is discovered at the lowest refinement level (in red), the entire check terminates (the last sphere in grey is skipped).
	}\label{fig:rake}
\end{figure}

\sbmps validate not only individual states, but also the motions between them. %
Typically, motion validation requires discretizing a continuous motion and validating each state in the discretization.
Thus, this is where \sbmps spend most of their time, and a significant body of work~\cite{Bohlin2000, Mandalika2019, Strub2022} has gone into reducing the number of edges validated during planning.
However, here is where our perspective on vector-lifted primitives shines: by combining our previously developed insights into vectorized \fk and individual state \cc, we unearth further algorithmic insights to improve the performance of motion validation via vectorization.

For efficient motion validation, we want to stop checking invalid motions as quickly as possible.
Assuming uniform probability of collision along a motion\footnote{In reality, for, \eg motion toward objects, this distribution is not uniform.}, \(\frac{n}{2}\) \cc attempts are wasted (in expectation) for an invalid motion discretized into \(n\) states.
Due to our perspective on vector-lifted \fk and \cc, we can reduce the amount of wasted computation by testing a \emph{spatially distributed} set of states in parallel.
Without loss of generality, consider a vector of eight states, and a motion discretized into \(n\) states. %
We can \emph{simultaneously} check states \([0, \frac{n}{8}, \ldots, \frac{7n}{8}]\) for the cost of a single check\footnote{This is a slight simplification; vector operations have low but nonzero overhead, and using them as we do may prevent auto-vectorization.}, and---if no collisions are found---comb through the remaining states by incrementing each index, for at most \(\frac{n}{8}\) iterations.
We refer to this spatially distributed collision check as the "rake" (\cref{fig:rake}).
Beyond improving \cc throughput by decreasing the total number of checks required by a factor of the width of the vector, by spatially distributing the states checked, we increase the probability of exiting \cc early for invalid motions---the validity of close states is correlated, so we have a higher chance of finding an invalid state by testing along the entire motion at once, compared to, \eg the first eight states at once.
Other work has investigated spatially distributed collision check scheduling in both hardware~\cite{shah_energyefficient_realtime_2023} and non-parallel software~\cite{Bohlin2000}.

Although the pure narrowphase approach is highly effective, we augment it with a "mid-phase" check.
Specifically: we employ a hierarchy of increasingly refined sphere collision models of the robot, and use coarse levels of this hierarchy to avoid expensive checks at the finer levels.
We generate a sphere model for the robot with a single sphere per link, conservatively over-approximating the actual collision geometries.
If this sphere does not collide with a given obstacle, we know that the actual collision geometry cannot collide with that obstacle, and can skip checking the spheres of the higher-fidelity model.
Notably, this does not require the typical branching-heavy approaches to broadphase collision detection, \eg bounding volume hierarchies---our approach does not require or use the typical recursive tree structure or any update operations beyond \fk.

For efficiency, it is preferable to not compute poses for any of a robot's links that come \emph{after} a link in collision.
We exploit a property of our tracing compiler, which can re-order instructions topologically, %
to \emph{interleave} each sphere's collision check (environment and self-collision) within the generated \fk{} code, placing checks immediately after the position of the sphere has been computed, wasting almost no effort on irrelevant \fk{} computation and achieving a significant performance gain.
This interleaving is compatible with the previously-described hierarchical sphere tree.

\subsection{Bringing it Together: Design of the Planner}

We also leverage \simd instructions elsewhere to improve planner performance.
Although not required in general, we assume the configuration space of the robot is Euclidean and thus linear interpolation between two \aos configurations becomes simply adding and multiplying their vectors together.
Similarly, the $\ell_2$-norm is computed efficiently as a horizontal summation.
We use these improvements to quickly compute the intermediate configurations used in the rake as well as distances in our \nn data-structure~\cite{Ichnowski2019, Ichnowski2020}.

We have implemented two \sbmps: \rrtconnect~\cite{Kuffner2000} and \prm~\cite{Kavraki1996}, without algorithmic changes or additional complexity due to our focus on planner primitives.%
We have also implemented simplification algorithms: randomized shortcutting~\cite{Geraerts2007, Hauser2010} and B-spline smoothing~\cite{pan2012collision}.

%% file: sections/experiments.tex
\begin{figure}[t!]
	\includeinkscape[width=\linewidth]{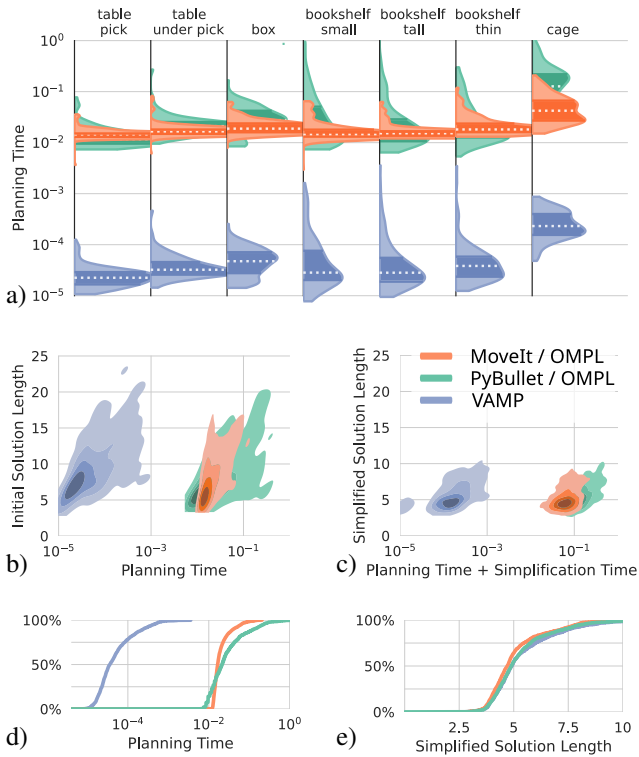}
	\caption{Results for the 7~\dof Panda.
   \textbf{a)} Planning times for each problem class.
   \textbf{b)} Planning time vs. initial path length and
   \textbf{c)} planning and simplification time vs. simplified path length for entire dataset.
   \textbf{d)} Cumulative distribution of planning time and
   \textbf{e)} cumulative distribution of simplified path length for entire dataset.
   All times are on a \textbf{logarithmic} scale.
	}
	\label{fig:panda}
\end{figure}

\begin{figure}[t!]
	\includeinkscape[width=\linewidth]{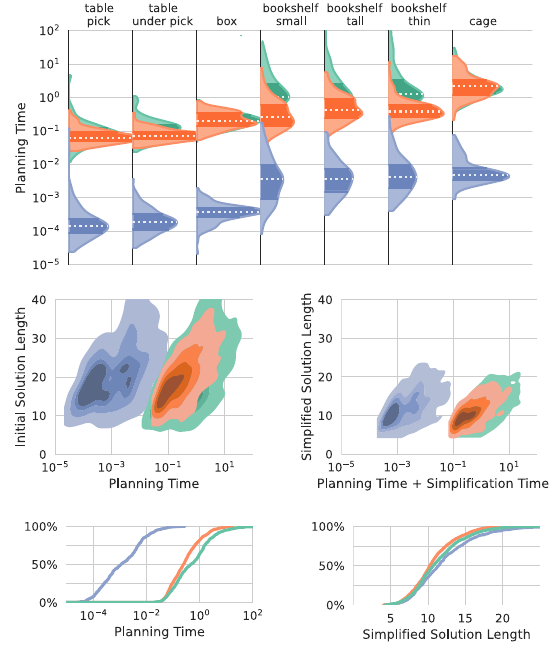}
	\caption{Results for the 8~\dof Fetch.
    \textbf{a)} Planning times for each problem class.
    \textbf{b)} Planning time vs. initial path length and
    \textbf{c)} planning and simplification time vs. simplified path length for entire dataset.
    \textbf{d)} Cumulative distribution of planning time and
    \textbf{e)} cumulative distribution of simplified path length for entire dataset.
    All times are on a \textbf{logarithmic} scale.
  }
	\label{fig:fetch}
\end{figure}

\begin{figure}
	\includeinkscape[width=\linewidth]{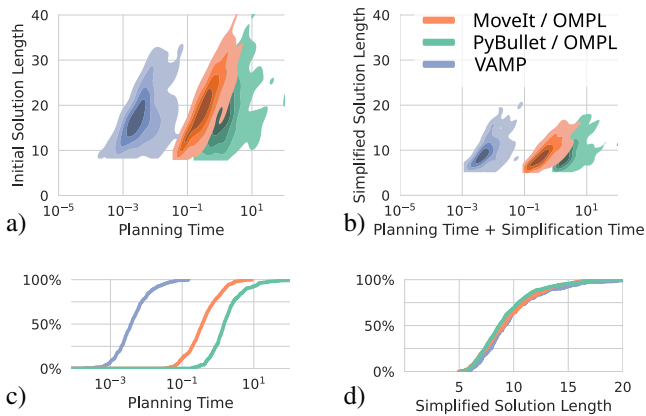}
	\caption{Results for the 14~\dof Baxter over entire dataset.
    \textbf{a)} Planning time vs. initial path length.
    \textbf{b)} Planning time plus simplification time vs. simplified path length.
    \textbf{d)} Cumulative distribution of planning time and
    \textbf{e)} cumulative distribution of simplified path length.
    All times are on a \textbf{logarithmic} scale.
	}
	\label{fig:baxter}
\end{figure}

\begin{table}
  \centering
  \tiny
  \begin{tabularx}{\linewidth}{ r | r | r | r r r r | r | r}
    & & & & & & & \multirow{2}{*}{\shortstack[c]{Mean\\Simpl.}} & \\
     & System & Mean & Q1 & Median & Q3 & 95\% & & Succ. \\
    \hline
    \hline
    \parbox[t]{0.1mm}{\multirow{4}{*}{\rotatebox[origin=c]{90}{Panda}}}
     & \grapeshot & 58.47 &    12.76 &    21.21 &    43.35 &   224.41 & 121.38 & 99.6\% \\
     & \moveit    & 23.51 &    14.28 &    16.58 &    22.54 &    63.08 & 46.95 & \textbf{100\%} \\
     & \vamparm   &  0.70 &     0.18 &     0.29 &     0.62 &     2.73 &     1.03 & \textbf{100\%} \\
     & \vamp      & \textbf{0.10} &     \textbf{0.02} &      \textbf{0.04} & \textbf{0.08} &     \textbf{0.43}  & \textbf{0.12} & \textbf{100\%} \\
    \hline
    \parbox[t]{0.1mm}{\multirow{4}{*}{\rotatebox[origin=c]{90}{Fetch}}}
     & \grapeshot & 3035.83 &   135.88 &   461.48 &  1544.34 & 10358.44 & 225.17 & 99.8\% \\
     & \moveit    & 788.68 &    94.11 &   243.80 &   666.56 &  3107.80  & 137.14 & \textbf{100\%} \\
     & \vamparm   & 28.24 &     1.72 &     5.76 &    23.25 &   116.74 &     3.93 & 99.5\% \\
     & \vamp      & \textbf{6.25} &     \textbf{0.25} &     \textbf{1.12} &     \textbf{4.92} &    \textbf{25.16}    & \textbf{0.52} & 99.5\% \\
    \hline
    \parbox[t]{0.1mm}{\multirow{4}{*}{\rotatebox[origin=c]{90}{Baxter}}}
     & \grapeshot & 4309.19 &   881.31 &  1500.44 &  2711.61 & 13100.54 & 938.62 & 99.3\% \\
     & \moveit    & 668.80 &   192.23 &   362.34 &   757.69 &  2267.35  & 96.81 & \textbf{100\%} \\
     & \vamparm   & 32.54 &     8.02 &    13.93 &    25.89 &   104.57 &     9.77 & \textbf{100\%} \\ 
     & \vamp      & \textbf{6.68} &     \textbf{1.22} &     \textbf{2.32} &     \textbf{4.68} &    \textbf{22.97}    & \textbf{1.22} & \textbf{100\%} \\
  \end{tabularx}
  \caption{Planning time for \rrtconnect from~\mf{fig:panda,fig:fetch,fig:baxter}.
    The mean, first quantile, median, third quantile, and 95\% quantile are shown, along with mean simplification time and success rate.
    All times are in \textbf{milliseconds}.
  }
  \label{tbl:results}
\end{table}

\begin{table}
  \centering
  \tiny
  \begin{tabularx}{\linewidth}{ r | r | r | r r r r | r}
     & System & Mean & Q1 & Median & Q3 & 95\% & Succ. \\
    \hline
    \hline
    \parbox[t]{0.1mm}{\multirow{3}{*}{\rotatebox[origin=c]{90}{Panda}}}
    & \grapeshot & 4481.66 &   127.63 &   328.73 &  1328.37 & 14183.90 & 99.1\% \\
    & \moveit    & 615.58 &   415.14 &   416.73 &   418.23 &  1116.90 & 96.2\% \\
    & \vamp      & \textbf{11.12} &     \textbf{0.16} &     \textbf{0.37} &     \textbf{1.69} &    \textbf{39.39} & \textbf{99.7\%} \\
    \hline
    \parbox[t]{0.1mm}{\multirow{3}{*}{\rotatebox[origin=c]{90}{Fetch}}}
    & \grapeshot & 36422.87 &  2417.08 & 13060.40 & 38550.10 & 174517.60 & 71.3\% \\
    & \moveit    & 4514.73 &   468.71 &  1037.15 &  3000.64 & 23895.72 & 85.7\% \\
    & \vamp      &  \textbf{337.23} &     \textbf{7.69} &    \textbf{30.09} &   \textbf{181.73} &  \textbf{2014.40} & \textbf{94.7\%} \\
  \end{tabularx}
  \caption{Planning times for \prm over problem classes \emph{table pick}, \emph{table under pick}, and \emph{box}.
    The mean, first quantile, median, third quantile, 95\% quantile, and success rate  are shown.
    All times are in \textbf{milliseconds}.
  }
  \label{tbl:prm}
\end{table}

We evaluate our approach against two baselines which use the Open Motion Planning Library (\ompl)~\cite{Sucan2012}: MoveIt~\cite{chitta2012moveit} through Robowflex~\cite{kingston2022robowflex} (\moveit) and \ompl's Python bindings with PyBullet~\cite{Coumans2016} (\grapeshot).
These represent two common interfaces of motion planning in practice: the standard motion planner for \ros~\cite{Quigley2009}, and a Python implementation using a popular simulation framework.
We evaluate our implementation, "Vector Accelerated Motion Planning" on an x86-based desktop computer (\vamp) as well as a small \arm-based single-board computer (\vamparm)\footnote{For \vamp, \avxtwo was used. The authors attempted using \avxthree, but found lower throughput than \avxtwo, possibly due to downclocking, lack of 512-bit registers, or other issues that will be investigated in future work. For \vamparm, \arm's Neon \simd instructions were used.}.
All hyperparameters are shared between each implementation: all planners use equivalent implementations of algorithms with identical validity checking resolution.
Moreover, we determinize all planners by sampling from a multi-dimensional Halton sequence~\cite{Halton1960, LaValle2004, Hsu2007}.
The same sequence is used between all systems.
Thus, performance differences can be attributed to vector-acceleration\footnote{\moveit uses the $\ell_1$ metric for nearest neighbors rather than $\ell_2$.}.

All benchmarks for \moveit, \grapeshot, and \vamp were performed with a \amd Ryzen\texttrademark~9 7950X \cpu clocked at 4.5GHz.
For \vamparm, benchmarks were run on an Orange Pi 5B with an \arm Cortex-A76 \cpu clocked at 2.4GHz.
Our approach is implemented in C++17 with Python bindings through \texttt{nanobind}~\cite{Jakob2022}.
All code (including \ompl and MoveIt) was compiled using \texttt{clang 15.0.7} with the \texttt{-Ofast} optimization level\footnote{Note that some of the issues with \texttt{-ffast-math}, \eg handling non-finite values, subnormals, etc., are not particularly relevant for the motion planning case, where configurations are from a compact, closed, and bounded space with relatively similar range in each dimension. However, we warn practitioners to still be wary of issues arising from reciprocal approximation.} and with all architecture optimizations (\texttt{-march=native}, \emph{i.e.}, \texttt{znver4}).

We evaluate on seven different environments from the MotionBenchMaker~\cite{chamzas2021mbm} dataset, a collection of realistic, difficult motion planning problems:
\begin{enumerate*}
	\item \emph{table pick} and \emph{table under pick} environments to evaluate tabletop manipulation,
	\item \emph{bookshelf small}, \emph{tall}, and \emph{thin} to demonstrate reaching, and
	\item \emph{box} and \emph{cage} to demonstrate highly constrained reaching.
\end{enumerate*}
We use the publicly available pre-generated 100 problems for each of the environments.
We evaluate on the following systems:
\begin{enumerate*}
\item the 7~\dof Franka Emika Panda\footnote{We use the approximation of the Panda from~\citet{fishman2023mpinets}},
\item the 8~\dof Fetch Robotics Fetch, including the prismatic torso joint, and
\item the 14~\dof bimanual Rethink Robotics Baxter
\end{enumerate*}.
 For the Baxter, we use the \emph{bookshelf tall \{easy, medium, hard\}} datasets for bimanual manipulation.
For Fetch and Baxter, we use the algorithm of~\citet{Bradshaw2004} to automatically generate a spherized model (after ensuring manifold meshes~\cite{Huang2018, Huang2020})---these approximations were also manually tuned.
We evaluate each planner on each problem 5 times.
For \moveit and \grapeshot, we give a timeout of 5 minutes.
For \vamp and \vamparm, we give a limit of 1 million planner iterations.

Results for \rrtconnect on the Panda, Fetch, and Baxter are respectively shown in~\mf{fig:panda,fig:fetch,fig:baxter} and summarized in~\mf{tbl:results}.
We note the following general features of these plots:
\begin{enumerate*}
\item times are all reported on a logarithmic scale,
\item the distribution of planning time for \vamp is almost completely separated from both \grapeshot and \moveit, and
\item the distribution shapes of planning time versus path length are qualitatively similar and simplified path length distributions are equivalent for each planner, indicating planner similarity at the algorithmic level
\end{enumerate*}.
Over all robots, \vamp is roughly 500x faster than \grapeshot and 100 to 200x faster than \moveit, while achieving similar path quality.
\vamp provides high-quality plans at control frequencies, \eg \SI{10}{\kilo\hertz} mean, \SI{25}{\kilo\hertz} median, and \SI{2.3}{\kilo\hertz} 95\% planning rates for the Panda arm for the entire dataset, which includes trivial problems such as tabletop manipulation and complex problems such as reaching into shelves.
Note that even the slowest number in~\mf{tbl:results}, \SI{25}{\milli\second} for the Fetch's 95\% quantile, achieves a \SI{40}{\hertz} planning rate. %
Morever, \vamparm also achieves similar speed-ups on a low-power single-board computer (the Orange Pi 5B uses up to \SI{7}{\watt}), still 20--50x faster than baselines on a desktop \cpu.
We also report times for \prm for the Panda and Fetch over the \emph{table pick}, \emph{table under pick}, and \emph{box} environments (\mf{tbl:prm}), and show the same caliber of performance improvements, indicating our approach generalizes across \sbmps.

%% file: sections/discussion.tex
Efficient motion planning is critical for many applications of robotics.
In this paper, we demonstrate a novel approach to accelerating motion planning, based on a new perspective on \emph{vector-oriented} operations for simple, high-frequency interleaving of high-performance parallelized and serial sections of code present in most sampling-based motion planning algorithms.
By applying this perspective to the most expensive and ubiquitous motion planning subroutines (\ie collision checking, forward kinematics, and distance computation), we achieve algorithmic improvements and create proof-of-concept planners that achieve more than 500x speedup over the state of the art on realistic, challenging planning problems for three different robots.
Our approach solves planning problems at kilohertz rates on ordinary consumer CPUs and low-power single-board computers.

We also believe that our ideas will extend naturally to harder motion planning problems, such as kinodynamic and manifold-constrained planning.
Further, because we can produce so many motion plans so fast, we may be able to efficiently provide empirical proofs of \emph{solution nonexistence}, a feat that has long been challenging for \sbmp.
There may also be potential for using our vector-oriented planners as \emph{local planners} inside higher-level motion planning algorithms.

The planning performance demonstrated in this work pushes \sbmp{} for high-DoF manipulators to frequencies required for control---providing a complete, global, high-quality plan at each update.
We believe that this is cause to re-examine old assumptions in robotics about the roles of planning and control, as well as about the "best" way to solve problems such as planning under uncertainty or integrated task and motion planning.
In particular, we are excited to explore extensions of this work around, \eg rapid replanning, integrated task and motion planning, etc.
In general, there is a rich discussion to be had about implications for algorithms that use motion planning as a subroutine, and that have traditionally needed to be designed around motion planning as an \emph{expensive} subroutine, now that we can consistently provide high-quality motion plans at high frequencies.